\documentclass[runningheads]{llncs}
\usepackage[T1]{fontenc}
\usepackage[utf8]{inputenc}
\usepackage{graphicx}
\usepackage{url}

\title{Sanyu Studio: A Multi-Agent System for Art-Historical Narrative Construction}
\titlerunning{Sanyu Studio}
\author{Zhaoxi Wei\inst{1} \and Hongye Yang\inst{2} \and Shuyuan Tian\inst{3}}
\authorrunning{Z. Wei et al.}
\institute{Renmin University of China\\\email{weizhaoxi25@ruc.edu.cn}
\and Georgia Institute of Technology\\\email{hyang783@gatech.edu}
\and University of Cologne (Universit\"at zu K\"oln)\\\email{stian@smail.uni-koeln.de}}

\begin{document}
\maketitle
\begin{abstract}
Generative AI offers new ways to explore art history, but its responses can also make different interpretations sound alike. We present Sanyu Studio, an interactive multi-agent system that helps users build and revise narratives about the artist Sanyu. The system represents 321 oil paintings as agents. Each agent draws on artwork facts and related interpretations, while a document-scarcity mechanism gives less-studied works a chance to contribute. Users question these agents in a roundtable, and a Sanyu agent uses selected memories to form a changing account of the artist. We studied the system in a seven-day workshop with eight art students. Although participants began with the same evidence, their questions and paths through the system led to different final narratives. These observations show how a multi-agent interface can make evidence selection and interpretation visible to users. They also highlight the need to distinguish documented facts from speculative claims.
\keywords{Multi-agent systems \and Art history \and Human--AI interaction \and Digital cultural heritage}
\end{abstract}

\section{Introduction}
Art-historical narratives grow from artworks, archives, and critical writing. Researchers select and connect this evidence, and often disagree about its meaning. Large language models (LLMs) can help people explore such sources, but may also narrow the range of ideas they encounter~\cite{anderson}. Can an AI interface instead support different, evidence-aware accounts of the same artist?

Multimodal models describe paintings~\cite{gallery}; retrieval systems connect art with external sources~\cite{artrag}; and historical chatbots support cultural learning~\cite{dacosta}. Yet users rarely see how several works offer distinct positions, or how their own questions shape a longer account.

We focus on Sanyu, a Chinese-born painter who worked in Paris. Parts of his life remain open to interpretation~\cite{yan}. Sanyu Studio lets users ask questions, gather works, compare evidence, retain interpretations, and revisit a narrative. We describe the system and a workshop with eight students to examine how shared evidence can lead to different stories.

\begin{figure}[t]
\centering
\includegraphics[width=.875\textwidth]{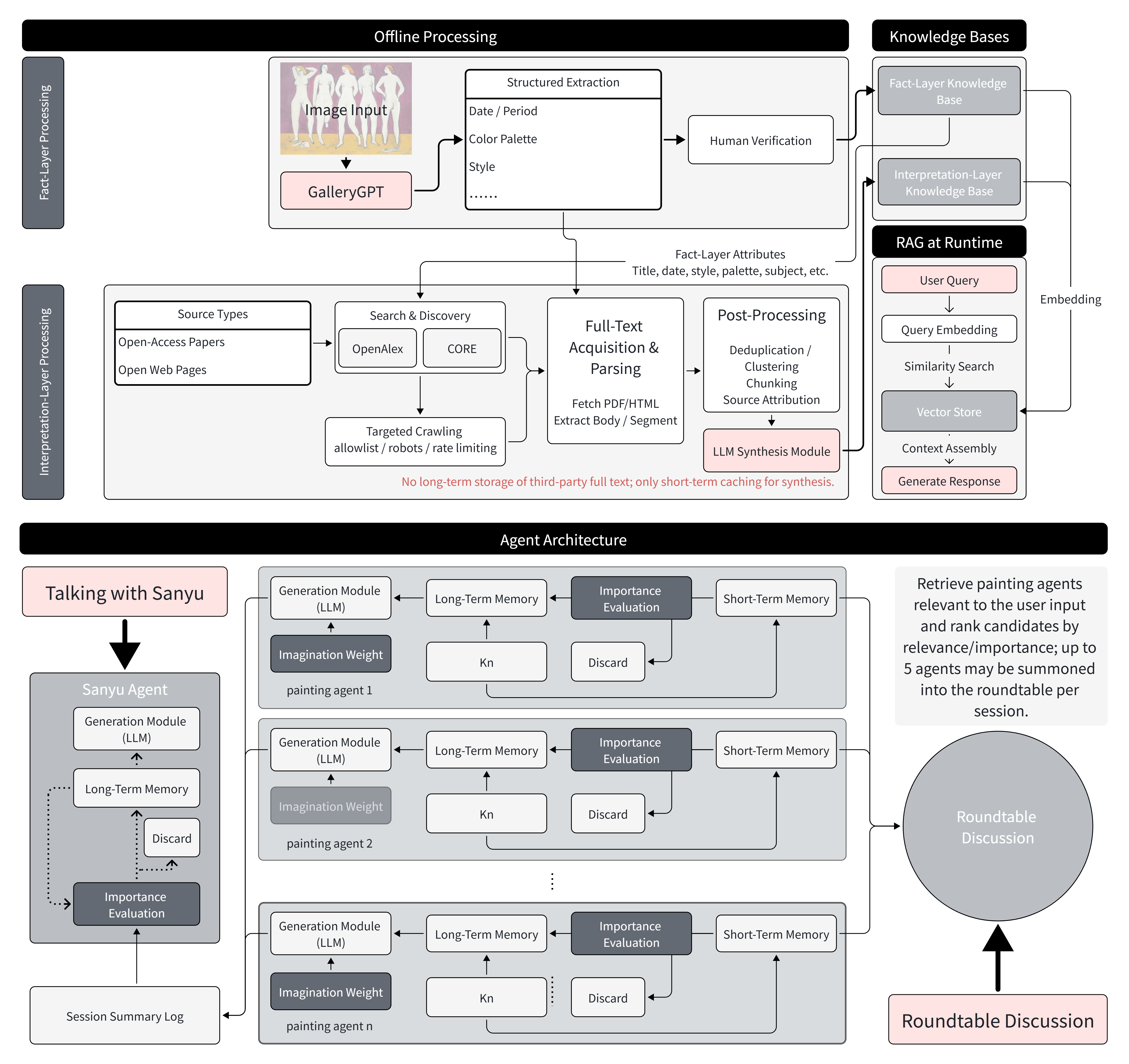}
\caption{System architecture. Artwork facts and scholarly interpretations are prepared as separate knowledge layers. Painting agents use these layers during the roundtable, while the Sanyu agent keeps selected knowledge across sessions.}
\label{fig:architecture}
\end{figure}

\section{System Design}
A user asks about Sanyu, relevant painting agents join a roundtable, and the system selects useful material from their exchange. A Sanyu agent combines these memories into a narrative that can be questioned and revised. Figure~\ref{fig:architecture} shows the evidence pipeline and agents.

\subsection{Painting Agents and Knowledge Base}
The system models 321 Sanyu oil paintings as agents, each speaking from its own evidence. A factual layer stores catalogue details, exhibition records, and visual descriptions supported by GalleryGPT~\cite{gallery}. An interpretation layer stores short, attributed entries from criticism and scholarship. Entries were manually checked against their sources; retrieval uses compressed entries and source metadata.

For each candidate agent, the system retrieves its five most relevant entries. Facts come from the factual layer, while uncertain interpretations are marked. This allows a painting to offer a viewpoint without presenting speculation as historical fact.

\begin{figure}[t]
\centering
\includegraphics[width=.935\textwidth]{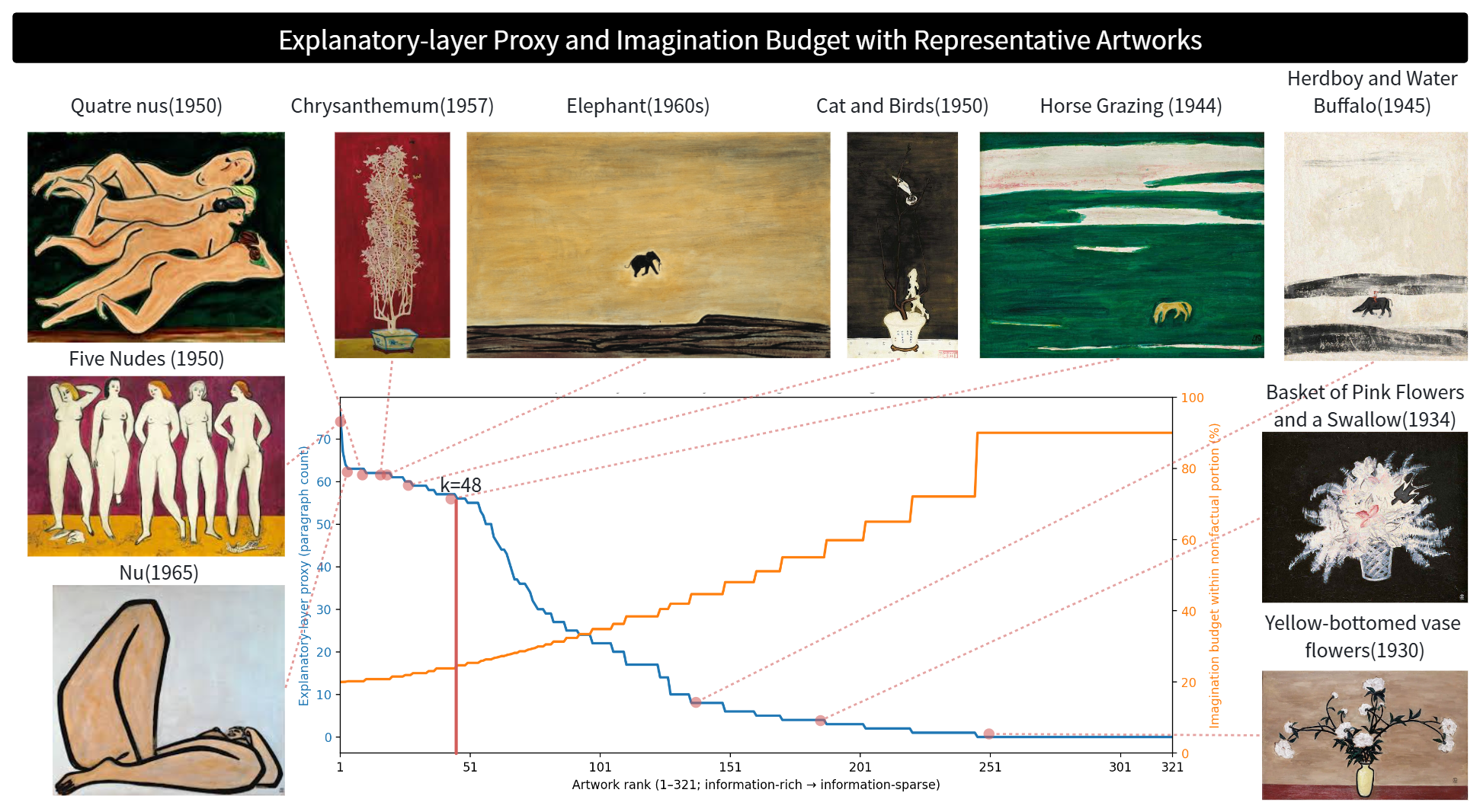}
\caption{Uneven coverage across 321 paintings. The blue curve measures available interpretive passages; the orange curve shows the DSBM inference budget. The first 48 paintings account for about half of the interpretive text.}
\label{fig:coverage}
\end{figure}

\subsection{Document-Scarcity Balancing Mechanism}
Documentation is uneven: the 48 most discussed paintings account for about half of the deduplicated interpretive passages across all 321 works. This raises a depth--coverage concern also seen in CAD auditing~\cite{depthbench}.

To broaden the discussion, we use a document-scarcity balancing mechanism (DSBM). It ranks candidate works using five signals: relevance to the question, lack of documentation, difference from already selected works, an evidence gate, and the share of earlier speaking time. Up to five painting agents take part in one roundtable. Scarce documentation can raise a work's chance of selection, but it does not relax the rules for factual claims. Where evidence is thin, an agent can suggest a clearly labeled interpretation based on visual comparison~\cite{ginzburg}.

In comparison, DSBM reduced the speaking share of high-information works from 41.8 to 20.4 percent. The number of distinct works discussed rose from 151 to 279. This suggests that balancing changes the evidence users encounter, beyond the wording of an answer.

\subsection{Memory and Interaction}
Each painting agent keeps recent dialogue in short-term memory and stores up to 100 selected fragments in long-term memory. After a roundtable, the system scores possible memory fragments from 1 to 10 for relevance, new evidence, and links to earlier discussion. Scores of at least 7 are retained. Older entries can be merged when the memory budget is full. The factual base remains fixed; memories track interpretation and may need updating, as with historical action credit~\cite{credit}.

\begin{figure}[t]
\centering
\includegraphics[width=.905\textwidth]{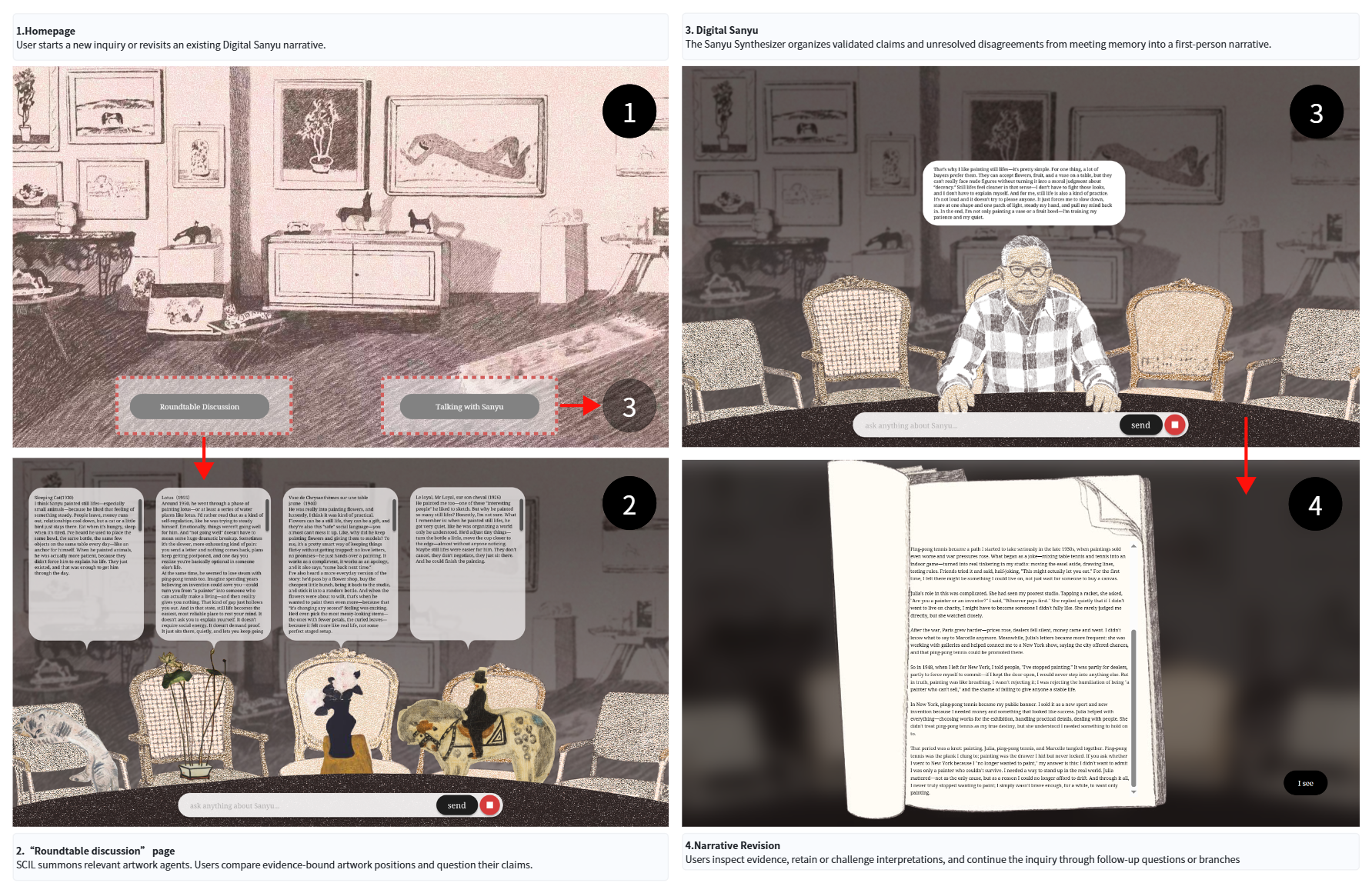}
\caption{Four stages of interaction: (1) start an inquiry, (2) compare painting agents in a roundtable, (3) talk to the Sanyu agent, and (4) read and revise the developing narrative.}
\label{fig:interaction}
\end{figure}

The Sanyu agent receives summaries from the painting agents and builds a longer account across sessions. After ten roundtables, users can talk directly with this agent. They can also return to the paintings and challenge earlier interpretations. Figure~\ref{fig:interaction} shows the four main views: the entry page, the roundtable, dialogue with Sanyu, and the revised narrative. The interface makes the process of writing an artist's story open to further questions.

\section{Workshop}
\subsection{Procedure}
We ran an exploratory, seven-day workshop with eight senior oil-painting students at an art university in China. Four were men and four were women; their mean age was 22 years (SD 1.20). Participants gave informed consent for the study and anonymous reporting. Each started with the same evidence base and system state. After a short introduction, they explored Sanyu Studio freely and completed at least 50 roundtable sessions.

Five shared questions covered Sanyu's friends in Paris, the care of his works, Henri-Pierre Roch\'e, mirror paintings, and ``ping-pong tennis.'' These topics have incomplete or debated records. We saved dialogue paths, retrieved evidence, and agent memories. Participants completed the brief Need for Closure Scale (NFCS)~\cite{roets}; scores ranged from 29 to 68. We treat these data descriptively.

\begin{figure}[t]
\centering
\includegraphics[width=.985\textwidth]{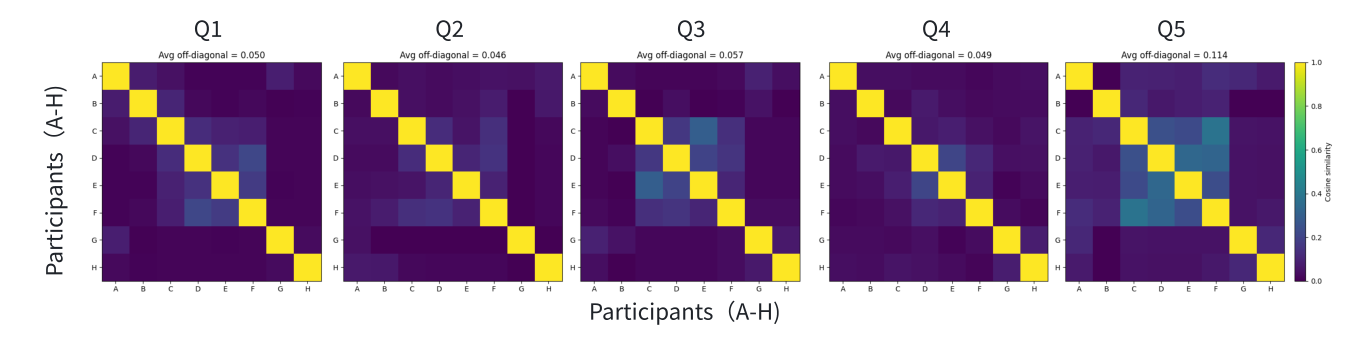}
\caption{Pairwise similarity of final answers from participants A--H to five shared questions. Bright diagonals compare an answer with itself; most off-diagonal similarities are low. Scores use cosine similarity after TF--IDF and latent semantic analysis.}
\label{fig:results}
\end{figure}

\subsection{Results}
At the end of the workshop, we asked each participant's Sanyu agent to answer the same five questions. We then compared answers using cosine similarity in a TF--IDF-based latent semantic analysis space. The mean pairwise similarity across different participants was 0.050, 0.046, 0.057, 0.049, and 0.114 for Questions 1--5, respectively (Fig.~\ref{fig:results}). The low scores show that the generated wording and themes differed, even when the final prompts were identical. They do not, on their own, measure the historical accuracy or quality of those answers.

Interaction logs offered a second pattern. Students with higher NFCS scores tended to stay with one topic, ask shorter questions, and return to the same line of inquiry. Those with lower scores more often switched topics or revisited earlier questions. For example, participant D (NFCS 68) held 53 sessions with a topic-switch rate of 0.286, while participant F (NFCS 29) held 72 sessions with a rate of 0.648. These are descriptive associations; we cannot assign causality to NFCS in this small group.

\section{Discussion and Limitations}
The workshop suggests two useful roles for Sanyu Studio. First, the roundtable makes the choice of evidence visible. Users see that paintings can support different aspects of a question, including those rarely discussed in writing. Second, memory lets a conversation become an evolving account. This follows an important feature of historical writing: a narrative is shaped by what is selected, connected, and remembered~\cite{ginzburg}.

There are important limits. Diverse answers do not ensure reliable interpretations, a concern shared with tool-agent evaluation~\cite{parthack}. The interface separates facts from conjecture, but agents can still infer too much about private experiences or relationships for which the record is weak. Stronger source checks, clearer uncertainty warnings, and human review are needed. Our sample also consisted of only eight art students. It was too small and too specialized to establish effects on learning or to generalize to visitors with little art-history experience. A future study should include a larger and more varied audience, direct checks of factual accuracy, and comparisons with simpler retrieval interfaces.

DSBM brings under-documented works into view, but some may invite unsupported claims. Source checks, uncertainty labels, and expert assessment remain essential. Memory also makes the narrative depend on earlier questions. Our workshop did not isolate how much of the final variation came from memory, agent selection, or user behavior. Controlled comparisons with those components disabled could help answer this question.

\section{Conclusion}
Sanyu Studio treats art-historical writing as a process that users can enter and revise. Its 321 painting agents provide distinct sources, DSBM helps less-studied works participate, and the Sanyu agent brings selected discussions into a developing narrative. In a seven-day workshop, users reached different accounts from shared starting evidence. These findings support further study of multi-agent interfaces for cultural heritage, especially where sources are incomplete and interpretations remain open.

{

}
\end{document}